\pdfoutput=1
\documentclass[journal,9pt]{IEEEtran}
\usepackage[latin9]{inputenc}
\usepackage{graphicx}
\usepackage{etoolbox}
\usepackage{amsfonts}
\usepackage{amssymb}
\usepackage{amsmath}
\usepackage{enumitem}
\usepackage{caption}
\usepackage{mathenv}
\usepackage{fancyhdr}
\usepackage{bm}
\usepackage{mathrsfs}
\usepackage{graphicx,color}
\usepackage{amssymb, amsmath, array}
\usepackage{hyperref}
\usepackage{float}
\usepackage{subfig}
\usepackage{verbatim}
\usepackage{harmony}
\usepackage{subfig}

\makeatletter
\ifCLASSINFOpdf
\else
\fi
\makeatother

\begin{document}
% paper title
\title{Conditional LSTM-GAN for Melody Generation from Lyrics}
\author{\IEEEauthorblockN{Yi Yu$^{1}$, Abhishek Srivastava$^{2}$, Simon Canales$^{3}$}\thanks{\textbf{Abhishek was involved in this work during his internship from August to September, 2020 in National Institute of Informatics (NII), Tokyo. Simon was involved in this work during his internship from June to August, 2019 in National Institute of Informatics (NII), Tokyo}.},
\\
 \IEEEauthorblockA{ $^{1}$Digital Content and Media Sciences Research Division, National Institute of Informatics, Tokyo, Japan\\
 $^{2}$Multimodal Digital Media Analysis Lab, Indraprastha Institute of
Information Technology Delhi, India\\
 $^{3}$Institut de g\'enie \'electrique et \'electronique, \'Ecole Polytechnique F\'ed\'erale de Lausanne, Switzerland\\
 }}
\maketitle
\begin{abstract}
Melody generation from lyrics has been a challenging research issue in the field of artificial intelligence and music, which enables to learn and discover latent relationship between interesting lyrics and accompanying melody. Unfortunately, the limited availability of paired lyrics-melody dataset with alignment information has hindered the research progress. To address this problem, we create a large dataset consisting of 12,197 MIDI songs each with paired lyrics and melody alignment through leveraging different music sources where alignment relationship between syllables and music attributes is extracted. Most importantly, we propose a novel deep generative model, conditional Long Short-Term Memory - Generative Adversarial Network (LSTM-GAN) for melody generation from lyrics, which contains a deep LSTM generator and a deep LSTM discriminator both conditioned on lyrics. In particular, lyrics-conditioned melody and alignment relationship between syllables of given lyrics and notes of predicted melody are generated simultaneously. Extensive experimental results have proved the effectiveness of our proposed lyrics-to-melody generative model, where plausible and tuneful sequences can be inferred from lyrics.
\end{abstract}


\section{Introduction}
Music generation is also referred to as music composition with the process of creating or writing an original piece of music, which is one of human creative activities \cite{Wiggins}. Without understanding music rules and concepts well, creating pleasing sounds is impossible. To learn these kinds of rules and concepts such as mathematical relationships between notes, timing, and melody, the earliest study of various music computational techniques related to Artificial Intelligence (AI) has emerged for music composition in the middle of 1950s \cite{Hiller}. Markov models as a representative method of machine learning have been applied to algorithmic composition \cite{Ponsford}. However, due to the limited availability of paired lyrics-melody dataset with alignment information, research progress of lyrics-conditioned music generation has been obstructed.
%AI techniques have been widely exploited in music generation.

With the development of available lyrics and melody dataset and deep neural networks, musical knowledge mining between lyrics and melody has gradually become possible \cite{Jean},\cite{Yu}.
Melody \cite{melody} is a sequence of musical notes over time, in which each note is sounded with a particular pitch and duration. Generating a melody from lyrics is to predict a melodic sequence when given lyrics as a condition. Existing works, e.g., Markov models \cite{Marco}, random forests\cite{Margareta}, and recurrent neural network (RNN)\cite{Hangbo}, can generate lyrics-conditioned music melody. However, these methods cannot ensure that the distribution of generated data is consistent with that of real samples. Generative adversarial networks (GANs) proposed in \cite{Goodfellow} are a generative model which can generate data samples following a given distribution, and have achieved a great success in the generation of image, video, and text.

%Inspired by excellent improvements of text-to-video generation using GAN\cite{Yitong},
Inspired by the great success of GANs in various generative models in the area of computer vision and national language processing, we propose a conditional LSTM-GAN model to compose lyrics-conditioned melody where a discriminator can help to ensure that generated melodies have the same distribution as real ones. To the best of our knowledge, this is the first study that conditional LSTM-GAN is proposed for melody generation from lyrics, which takes lyrics as additional context to instruct deep LSTM-based generator network and deep LSTM-based discriminator network. Our proposed generation framework has several significant contributions, as follows:

i) A LSTM network is trained to learn a joint embedding in the syllable-level and word-level to capture syntactic structures of lyrics, which can represent semantic information of lyrics.

ii) A conditional LSTM-GAN is optimized to generate discrete-valued sequences of music data by introducing a quantizer.

%generate melody sequence using the encoded feature description.
iii) A large-scale paired lyrics-melody dataset with 12,197 MIDI songs is built to demonstrate that our proposed conditional LSTM-GAN can generate more pleasant and harmonious melody compared with baseline methods.

\section{Related works}

Automatic music generation has experienced a significant change in computational techniques related to artificial intelligence and music \cite{Jose}, spanning from knowledge-based or rule-based methods to deep learning methods. The majority of traditional music generation methods are based on music knowledge representation, which is a natural way to solve the issue with some kind of composition rules \cite{Anders}. Knowledge-based music rules are utilized to generate melodies when given the specified emotions by users \cite{Delgado}. Moreover, several statistical models \cite{Darrell} such as hidden Markov models, random walk, and stochastic sampling are discussed for music generation. For example, Jazz chord progressions are generated by Markov model for music generation in \cite{Arne} and statistical models are applied to music composition in \cite{Cope}.

With rapid advancement of neural networks, deep learning has been extended to the field of music generation. A hierarchical RNN for melody generation is proposed in \cite{Jian}, which includes three LSTM subnetworks. Beat profile and bar profile are exploited to represent rhythm features at two different time scales respectively. A neural network architecture \cite{Daniel} is suggested to compose polyphonic music with a manner of preserving translation invariance of dataset. Motivated by convolution that can obtain transposition-invariance and generate joint probability distributions over a musical sequence, two extended versions, Tied-parallel LSTM-NADE and bi-axial LSTM, are proposed to achieve better performance. A continuous RNN with adversarial training (C-RNN-GAN) \cite{Olof} is proposed to compose MIDI classical music. RNN is considered to model sequences of MIDI data during adversarial learning. The generator is to transform random noise to MIDI sequences, while the discriminator is to distinguish the generated MIDI sequence from real ones.

Earliest work \cite{Satoru} for lyrics-conditioned melody generation is defined as generating a melody when given Japanese lyrics, patterns of music rhythms, and harmony sequences. Some constraints are determined to associate syllables with notes. Melody generation is realized by dynamic programming. In \cite{Kristine} the rhythmic patterns occurred in notes can be classified. Pitches that are most suitable for accompanying the lyrics are generated using n-gram models. Three stylistic categories such as nursery rhymes, folk songs, and rock songs are generated for given lyrics. A recently proposed ALYSIA songwriting system \cite{Margareta} is a lyrics-conditioned melody generation system based on exploiting a random forest model, which can predict the pitch and rhythm of notes to determine the accompaniments for lyrics. When given Chinese lyrics, melody and exact alignment are predicted in a lyrics-conditional melody composition framework \cite{Hangbo}, which is an end-to-end neural network model including RNN-based lyrics encoder, RNN-based context melody encoder, and a hierarchical RNN decoder. The authors create large-scale Chinese language lyrics-melody dataset to evaluate the proposed learning model.

%The proposed model does not encode musical knowledge and rules.
Our work focuses on lyrics-conditioned melody generation using LSTM-based conditional GAN, which is first proposed for tackling melody generation from lyrics. Distincting from the existing works, our generative model can ensure that the distribution of generated melody mimics that of real sample melodies. A skip-gram model is trained to transform raw textual lyrics into syllable embedding vector which is taken as input together with noise vector for training a deep generator model. A deep discriminator model is trained to distinguish generated MIDI note sequences from real ones. A large English language lyrics-melody dataset is built to validate the effectiveness of our proposed recurrent conditional GAN for lyrics-to-melody generation. Novel ideas are designed to evaluate melody generation, which is also very useful reference for measuring various generative models.

\section{Preliminaries}
Before we describe our melody generation algorithm, a brief introduction is given in the following to help understanding musical knowledge, the sequential alignment relationship between lyrics and melody, and how to build the lyrics-melody music dataset.

\subsection{Melody}

Melody and lyrics provide complementary information in understanding a song with the richness of human beings' emotions, cultures, and activities. Melody, as a temporal sequence containing musical notes, plays an important role. A note contains two music attributes: pitch and duration. Pitches are perceptual properties of sounds that organize music by highness or lowness on a frequency-related scale, which can be played in patterns to create melodies \cite{pitch}. Piano keys have MIDI numbers ranging from 21 to 108, which also represent the corresponding pitch numbers. For example, `D5' and `A4' can be respectively represented as 74 and 69 according to the mapping between notes and MIDI numbers. In music, duration \cite{duration} represents the length of time that a pitch or tone is sounded. Rests \cite{rest} are intervals of silence in pieces of music, marked by symbols indicating the length of the pause. If we treat a rest as a note with a special pitch value, it is hard to find a syllable paired with it. Therefore, to ensure one-syllable-to-one-note alignment, in this work, we take the ``rest'' as one of music attributes contained in a sequence of triplets.

\subsection{Lyrics}
Lyrics as natural language represent music theme and story, which are a very important element for creating a meaningful impression of the music. An English syllable \cite{syllable} is a unit of sound, which may be a word or a part of a word. According to timestamp information in MIDI files of music tracks, melodies and lyrics are synchronized together to parse the data and extract alignment information.

\subsection{Alignment}
\begin{figure}[!h]
\centering \includegraphics[width=8.7cm]{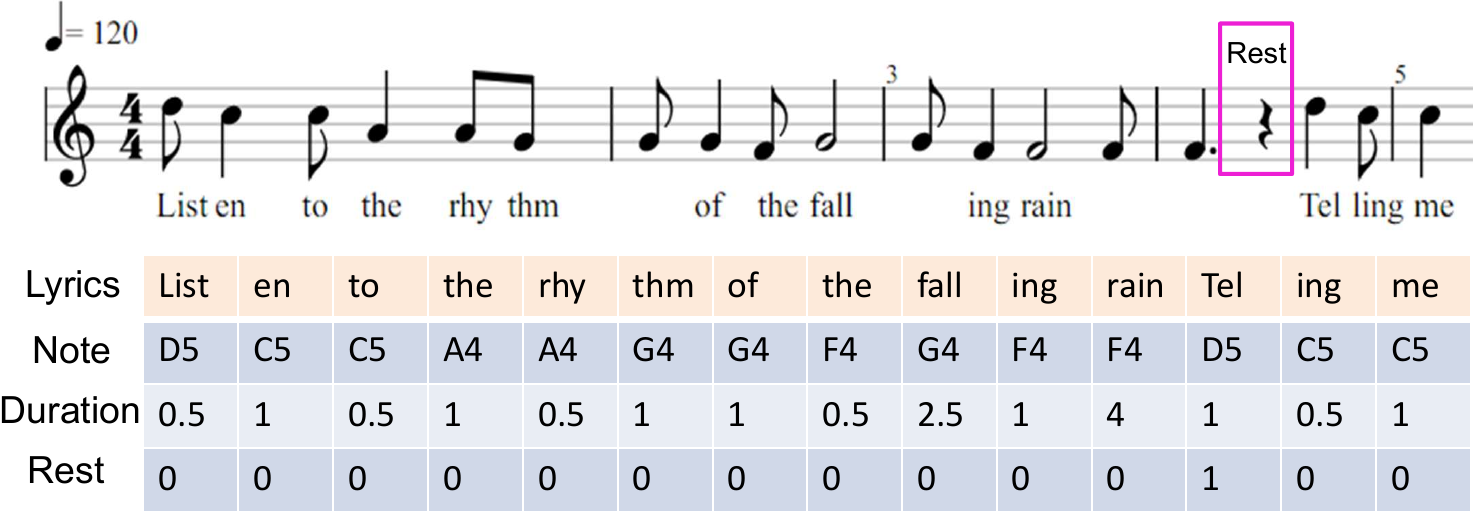} \caption{An example of alignment between lyrics and melody.}
\label{fig:alignment}
\end{figure}
An English syllable \cite{syllable} is a unit of sound, which may be a word or a part of a word. One syllable may correspond to one or more notes in English songs. Accordingly, the alignment between lyrics and melody could be a paired sequence of one-syllable-to-one-note or one-syllable-to-multiple-notes. Following the existing research such as \cite{Margareta} on melody generation from lyrics, currently we only exploit the pairs of one-syllable-to-one-note to train our conditional LSTM-GAN model for melody generation from lyrics, and investigate the pairs of one-syllable-to-multiple-notes in the future work. An example data structure of the alignment between lyrics and melodies is shown in Fig. \ref{fig:alignment}. Lyrics are divided to syllables. Each column represents one syllable with its corresponding triplet of music attributes \{note, duration, rest\}. Accordingly, a lyrics sample can be associated with a sequence of \{note, duration, rest\} triplets. With these music attributes, sheet music can be produced.

\section{Melody generation from lyrics}

Our aim is to learn a deep model that is able to represent the distribution of real samples, which further has the capability of generating new samples from this estimated distribution. In particular, using the capability of deep learning and generative modeling, sequential alignment relationship can be learned between lyrics and melody from real musical samples. As the number of epochs increases, the learning goes deeper, and the distribution of generated samples is more consistent with the distribution of real samples. Our training data contains the sequential alignment relationship between syllables and notes, which can be learned by conditional LSTM-GAN. During the training, GAN actually can resemble the distribution of the training samples and LSTM can learn the sequential alignment relationship.

\begin{figure}[!h]
\centering \includegraphics[width=8.8cm]{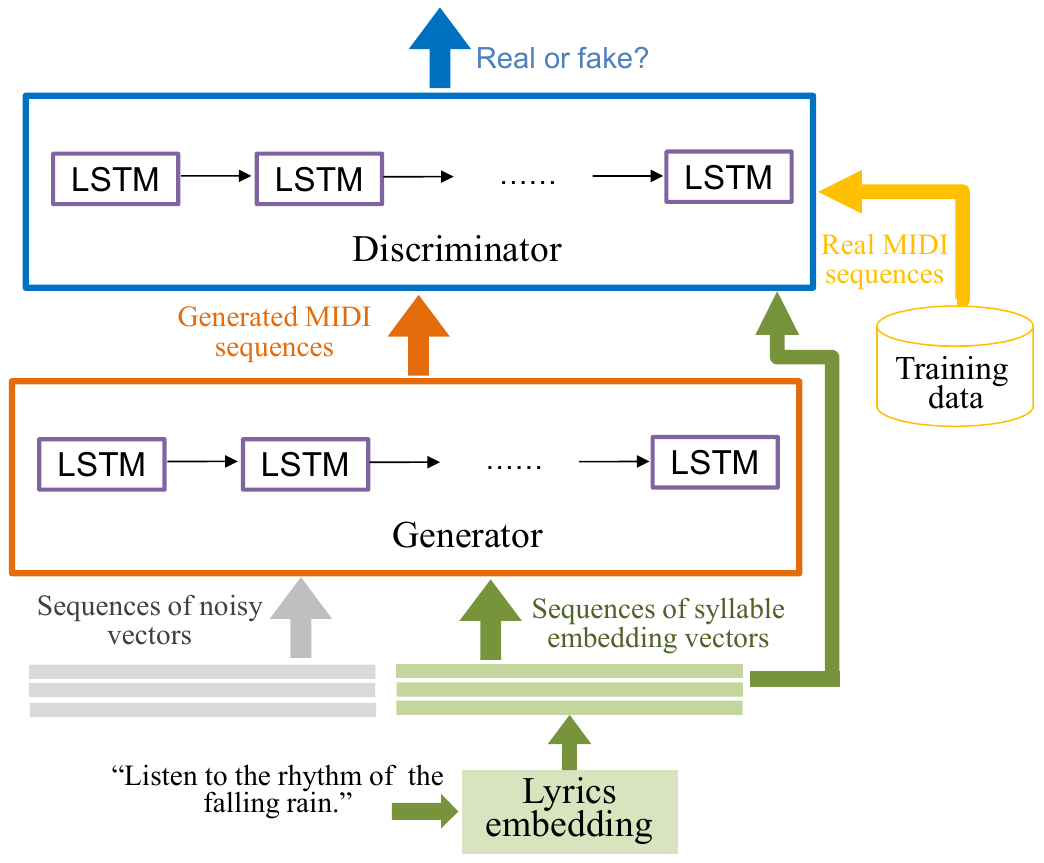} \caption{Conditional LSTM-GAN for melody generation from lyrics.}
\label{fig:framework}
\end{figure}

Our proposed conditional LSTM-GAN for lyrics-to-melody generation model is shown in Fig. \ref{fig:framework}, which is an end-to-end generative learning conditioned with lyrics. A sequence of syllable embedding vectors concatenated with noisy vectors is taken as input of the generator network. The generated MIDI sequences together with the sequence of syllable embedding vectors are taken as input of the discriminator network, which aims to train a model for distinguishing generated MIDI note sequences from real ones. In addition, a tuning scheme is introduced to quantize MIDI numbers so as to generate melody sequence with discrete attributes. Both the generator and discriminator are unidirectional RNN networks with the configuration shown in Table \ref{tab:configuration}.

\begin{table*}[!ht] \centering
% \resizebox{\linewidth}{!}
\begin{tabular}{|c|c|c|}
\hline
                 & RNN1 (generator)                             & RNN2 (discriminator)
\\ \hline
Input            & 30 (random noise), 20 (syll. embedding)     & 3 (MIDI attributes), 20 (syll. embedding) \\ \hline
Layer 1          & 400, Fully-connected, ReLU                  & 400, LSTM, tanh                            \\ \hline                        Layer 2          & 400, LSTM, tanh                             & 400, LSTM, tanh                            \\ \hline
Layer 3          & 400, LSTM, tanh                             & 2 (real or fake), sigmoid                            \\ \hline
Output & 3 (MIDI attributes), fully-connected, linear  & N/A                       \\ \hline
\end{tabular} %}
\caption{Configuration of the generator and discriminator.}
\label{tab:configuration}
\end{table*}

\subsection{Problem formulation}
Taking lyrics as input, our goal is to predict a melody sequentially aligned with the lyrics, where MIDI numbers, note duration, and rest duration are synthesized with lyrics to generate a song. Our research problem can be formulated as follows:
The syllables of lyrics as input are represented by a sequence $Y = (y^1, \cdots, y^{|K|})$. The melody as output is a sequence $X = (x^1, \cdots, x^{|K|})$, where MIDI numbers, note duration, and rest duration are simultaneously predicted as the $x^i = {\{x_{MIDI}^i, x_{dur}^i, x_{rest}^i}\}$. Moreover, the time length of the output sequence % $Y$ is constrained by the following condition:
\begin{equation}
\sum_{i=1}^{|K|} x_{dur}^i + x_{rest}^i  % |Y| =
\end{equation}
determines the length of synthesized song with lyrics.
% which also indicates that the generated melody should be exactly aligned with given lyrics.

\subsection{Lyrics embedding}

As the vocabulary of our lyrics data is large without any labels, we need to train an unsupervised model that can learn the context of any word or syllable. Specifically, we utilize our lyrics data to train a skip-gram model, which enables us to obtain the vector representation that can encode linguistic regularities and patterns in the context of given lyrics.

Our method encodes lyrics information at two different semantic levels: syllable-level and word-level, and two skip-gram models are trained respectively, which aim at associating a vector representation with an English syllable. Specifically,
lyrics of each song are divided into sentences, each sentence is divided into words, and each word is further divided into syllables. Words $W = \{w_1, w_2, w_3, ...,w_n\}$ are taken as tokens for training a word-level embedding model and syllables $S = \{s_1, s_2, ..., s_m\}$ are taken as tokens for training a syllable-level embedding model. Then, we train each skip-gram model as a logistic regression with stochastic gradient decent as the optimizer, and the learning rate with an initial value 0.03 is gradually decayed every epoch until 0.0007. Context window spans 7 adjacent tokens and negative sampling distribution parameter is $\alpha = 0.75$. We train the models to respectively obtain the word-level and syllable-level embedding vectors of dimensions $V = {10}$.

Let $\text{E}_{w}(\cdot)$ and $\text{E}_{s}(\cdot)$ denote the word-level and syllable-level encoders respectively, and $s$ denote a syllable
from word $w$. Then, syllable embedding and word embedding are concatenated as follows: % The concatenation of syllable $\text{E}_{s}(\cdot)$ and word $\text{E}_{w}(\cdot)$ is given by
\begin{equation}
\mathbf{y}=\text{E}_{w}(w) || \text{E}_{s}(s)=\mathbf{w} || \mathbf{s}
\end{equation}
where $\mathbf{s}  = \text{E}_{s}(s)\in\mathbb{R}^{10}$ and $\mathbf{w} = \text{E}_{w}(w) \in\mathbb{R}^{10}$ are the embedding of syllable $s$ and word $w$, respectively. The dimension of the overall embedding is $V = {20}$. Accordingly, lyrics sequences each with 20 syllables, each syllable being encoded in 20-dimension, are used in our experiment.

%\begin{figure}[!h]
%\centering \includegraphics[width=9cm]{images/wordlevelembedding}
%\caption{Word level encoding for the sentence "I love machine learning".}
%\label{fig:syll_enc}
%\end{figure}

\subsection{Condtional LSTM-GAN model}
In our work, an end-to-end deep generative model is proposed for lyrics-conditioned melody generation. LSTM is trained to learn semantic meaning and relationships between lyrics and melody sequences. Conditional GAN is trained to predict melody when given lyrics as input based on considering music alignment relationship between lyrics and melody.

\subsubsection{LSTM}
LSTM \cite{doi:10.1162/neco.1997.9.8.1735} networks are an extension to RNNs, which not only contain internal memory but also have capability of learning longer dependencies. An LSTM cell has three gates: input, forget, and output. These gates decide whether or not allow new input in, forget old information, and affect output at current time-step. In particular, at time-step $t$, the three states of the gates in an LSTM cell are given by:

\begin{equation}
    i_t = \sigma(w_i[h_{t-1},x_t] + b_i)
\end{equation}
\begin{equation}
    f_t = \sigma(w_f[h_{t-1},x_t] + b_f)
\end{equation}
\begin{equation}
    o_t = \sigma(w_o[h_{t-1},x_t] + b_o)
\end{equation}
where $i_t$, $f_t$ and $o_t$ denote the input, forget, and output gates states respectively, $h_{t-1}$ is the output of the LSTM cell at previous time-step, $w$'s and $b$'s are weights and biases, $x_t$ is the input of the LSTM cell, and $\sigma(\cdot)$ is the sigmoid function.

Then, the current output of the cell is computed by:
\begin{eqnarray}
    h_t &=& o_t \circ tanh(c_t) \\
    c_t &=& f_t\circ c_{t-1} + i_t \circ \Tilde{c}_t \\
    \Tilde{c}_t &=& tanh(w_c[h_{t-1},x_t]+b_c).
\end{eqnarray}
where $\circ$ denotes the element-wise multiplication between vectors.

\subsubsection{GAN}

A GAN is proposed by Ian Goodfellow, et al. \cite{Goodfellow}, which aims to train generative models by mitigating complex computation of approximating many probabilities. The general idea of GAN is to simultaneously train a generator $\text{G}(\cdot)$ and a discriminator $\text{D}(\cdot)$ with conflicting objectives. This method learns to predict new data with the same statistics as the training set. The generator $\text{G}(\cdot)$ tries to capture data distribution of training set. It takes an uniform noise vector $\mathbf{z}$ as an input and outputs a vector $\mathbf{\Tilde{x}}=\text{G}(\mathbf{z})$. In an adversarial way, the discriminator $\text{D}(\cdot)$  tries to identify samples produced by the generator from real ones. That is to say, $\text{G}(\cdot)$ and $\text{D}(\cdot)$ play the following two-player minimax game:
\begin{equation}
\begin{aligned}
\underset{G}{\operatorname{min}}\ \underset{D}{\operatorname{max}}\ V(D,G) &= \mathbb{E}_{\mathbf{x}\sim p_{\text{data}}(\mathbf{x})}[\text{log}D(\mathbf{x})] \\
 &+ \mathbb{E}_{\mathbf{z}\sim p_{\mathbf{z}}(\mathbf{z})}[\text{log}(1-D(G(\mathbf{z})))]
\end{aligned}
\end{equation}

%composed of two networks: a generator and a discriminator.
%The generator $\text{G}(\cdot)$ takes a noise
%vector $\mathbf{z}$ as an input and outputs a vector $\mathbf{\Tilde{x}}=\text{G}(\mathbf{z})$
%(note that $\mathbf{z}$ is a random variable and so is $\text{G}(\mathbf{z})$).
%The training aims at making sampling from G indistinguishable from
%"sampling" from the training data set. The generator is trained
%in an adversarial way, meaning that it competes with a discriminator
%$\text{D}(\cdot)$ which aims at recognizing whether a given vector
%$\mathbf{x}$ comes from the training data set or is produced by $\text{G}(\cdot)$.

\subsubsection{Lyrics-conditional GAN}
Conditional GAN is proposed in \cite{DBLP:journals/corr/MirzaO14}, with the goal of instructing the generation process by conditioning the model with additional information, which motivates us to train a generative model for lyrics-conditioned melody generation. In this work, the generator and discriminator are conditioned on lyrics. The lyrics are encoded to a sequence of 20-dimensional embedding vectors.

%GAN with class labels or data from different modality
%(in the present project text), it is possible to direct the generation process.
%Figure \ref{fig:cgan} illustrates the structure of a conditional GAN. \\

The melody contains a sequence
of music attributes $\mathbf{x}^{(i)}$, representing MIDI note, duration, and rest. Therefore, in the context of lyrics-conditioned melody generation, the input of the generator
is the paired sequences of syllable embedding vectors $\mathbf{y}^{(i)}$ and uniform random vector $\mathbf{n}^{(i)}$ in $[0,1]^{k}$, where $k=30$. The generated music attributes, syllable embedding vectors $\mathbf{y}^{(i)}$, and real music attributes $\mathbf{x}^{(i)}$, are fed to the discriminator.
$G(\mathbf{z}^{(i)}\lvert\mathbf{y}^{(i)})$ is a sequence of triplets containing attributes $\hat{x}^i = {\{\hat{x}_{MIDI}^i, \hat{x}_{dur}^i, \hat{x}_{rest}^i}\}$. Both the generator and the discriminator contain LSTM cells. The loss functions in the following are implemented to jointly train the generator and discriminator, where $m$ is mini batch size.
%where $k=30$ is the dimensionality of the data in the random sequence.
\begin{equation}
L_{G}=\frac{1}{m}\sum_{i=1}^{m}\text{log}(1-D(G(\mathbf{z}^{(i)}\lvert\mathbf{y}^{(i)})\lvert\mathbf{y}^{(i)} )).
\end{equation}

\begin{equation}
    L_{D}=\frac{1}{m}\sum_{i=1}^{m}[-\text{log}D(\mathbf{x}^{(i)}\lvert\mathbf{y}^{(i)})
    -\text{log}(1-D(G(\mathbf{z}^{(i)}\lvert\mathbf{y}^{(i)})\lvert\mathbf{y}^{(i)} ))].
\label{eq:loss_cond_d}
\end{equation}

\begin{figure}
\centering \includegraphics[width=8cm]{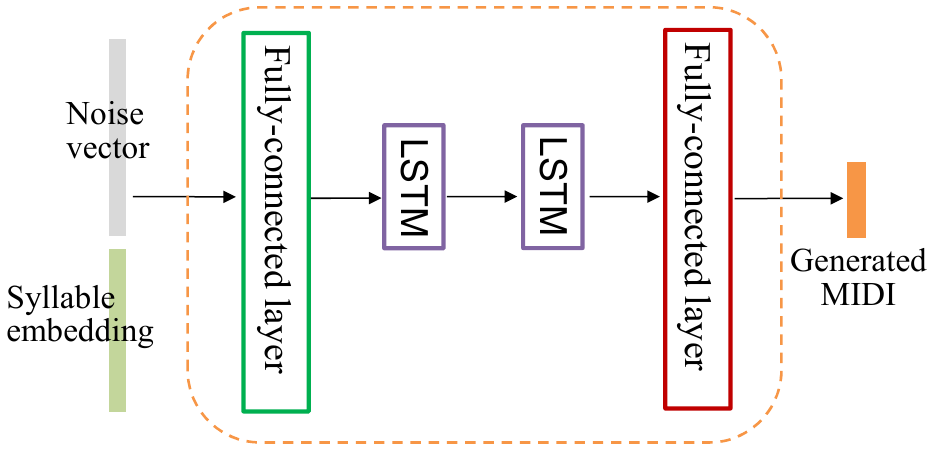} \caption{Generator network for one MIDI note generation, conditioned with an
encoded syllable $\mathbf{y}\in\mathbb{R}^{20}$, with an input random
noise vector $\mathbf{n}\in\mathbb{R}^{30}$, and output MIDI attributes
$\hat{x}\in\mathbb{R}^{3}$.}
\label{fig:gen_net}
\end{figure}

\subsubsection{Generator network in Fig.\ref{fig:gen_net}}

The generator is to learn the distribution of real samples, which is trained to increase the error rate of the discriminator. To learn a  distribution over the data, the generator builds a mapping function from a prior noise distribution to the data space. By concatenating the syllable embedding and noise as input, GAN is expected to instruct the melody  generation process (by lyrics) while generating diverse melodies (by random noise). In this work, each melody sequence has 20 notes, which needs 20 LSTM cells to learn the sequential alignment between lyrics and melody. The first layer in the generator network uses ReLU (rectified linear unit). When given a 50-dimensional vector concatenated by an encoded syllable $\mathbf{y}\in\mathbb{R}^{20}$ and an input random noise vector $\mathbf{n}\in\mathbb{R}^{30}$, the output of the first layer is scaled to a 400-dimensional vector to fit the number of internal hidden units of the LSTM cells. We tried different amounts of LSTM layers and found that 2 layers are sufficient in generator network for MIDI note generation. Then, the fourth linear layer produces a triplet of music attributes $\hat{x}^i = {\{\hat{x}_{MIDI}^i, \hat{x}_{dur}^i, \hat{x}_{rest}^i}\}$.

The same LSTM cell is used for each syllable in the lyrics. The output with the triplet music attributes of previous LSTM cell is concatenated with current 20-dimensional syllable embedding, which are further fed to current LSTM cell. This procedure is repeated until the generator can succeed to fool the discriminator.

%the same network of LSTM cell is used in series, and the states of the LSTM cell used for the preceding
%note is given to the next LSTM cell. \\

\begin{figure}
\centering \includegraphics[width=7.5cm]{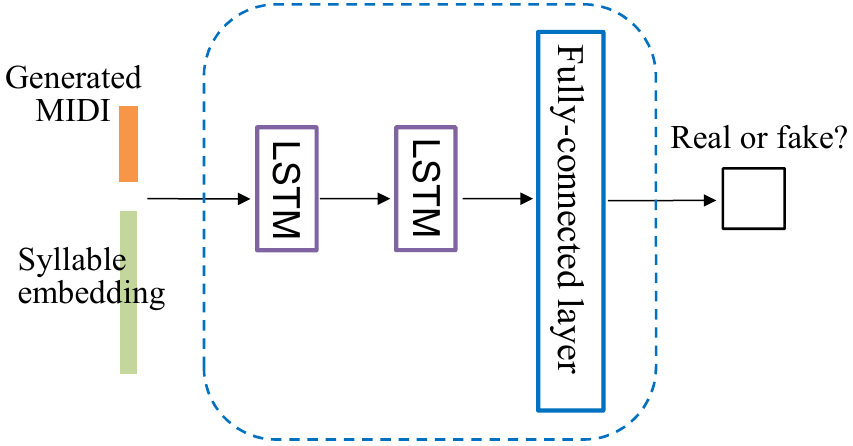} \caption{Discriminator network for one MIDI note, conditioned with an encoded syllable $\mathbf{y}\in\mathbb{R}^{20}$, with the generated MIDI attributes $\hat{x}\in\mathbb{R}^{3}$, and output the decision of real or fake.}
\label{fig:dis_net}
\end{figure}

\subsubsection{Discriminator network in Fig.\ref{fig:dis_net}}
%The discriminator consists of a two layers bi-directionnal LSTM cell,
The discriminator is to distinguish real melody samples from generated ones, which is trained by estimating the probability that a sample is from real training dataset rather than the generator. The discriminator loss function penalizes the discriminator for misestimating a real sample as fake or a fake sample as real. Because the input of the discriminator is a sequence, LSTM is also used here. Since lyrics as context information are used as condition in the discriminator, the generated triplet of music attributes concatenated with syllable embedding together as a 23-dimensional vector is input to the first LSTM cell in the discriminator. The hidden size of the LSTM cell in the discriminator is also 400. The output 400-dimensional vector from the second LSTM layer is input to the third linear layer, followed by a sigmoid activation function which estimates the decision output of real or fake by a value in the range $[0,1]$. With conditioning lyrics, the discriminator and generator are simultaneously learned until the training process converges.

\subsection{Tuning scheme}
The output from the generator is a continuous-valued sequence, which needs to be constrained to the underlying musical representation of discrete-valued MIDI attributes. Quantization of music attributes (MIDI number, note duration, and rest duration) are done during the generation in the experiments of validation and testing. Music attributes are constrained to their closest discrete values.

The concept of standard scales is important for the melody generation task. If a melody contains notes belonging to one of the standard scales, it indicates this melody has a perfect scale consistency. Below are three examples of standard scales \footnote{\url{https://en.wikipedia.org/wiki/Scale\_(music)}}:

\begin{itemize}
    \item $\mathcal{C}_{major}= \{C,D,E,F,G,A,B\}$;
    \item $\mathcal{D}_{major}= \{D,E,F\sharp,G,A,B,C\sharp\}$;
    \item $\mathcal{D}_{natural\ minor}= \{D,E,F,G,A,B\flat,C\}$, where $B\flat$ is equivalent to $A\sharp$.
\end{itemize}

The quantized music attributes are estimated to see if each generated sequence has a perfect scale consistency of melody. In particular, the most likely standard scale of music attributes this sequence belongs to is generated from syllable embedding in validation and testing datasets. The remaining out-of-tune notes are mapped to their closest in-tune music attributes.

\section{Lyrics-melody data acquirement}
There is no aligned lyrics-melody music dataset publicly available for music generation. In this work, a large-scale music dataset with sequential alignment between lyrics and melody is created to investigate the feasibility of this research with deep conditional LSTM-GAN. We acquire lyrics-melody paired data from two sources based on considering melodies with enough English lyrics, where 7,998 MIDI files come from the LMD-full MIDI Dataset \cite{lmd} and 4,199 MIDI files come from the reddit MIDI dataset \cite{reddit}. Altogether there are 12,197 MIDI files in the dataset, which contain 789.34 hours of melodies. The average length of melodies is 3.88 minutes. This dataset is available on Github  \footnote{\url{https://github.com/yy1lab/Lyrics-Conditioned-Neural-Melody-Generation}}.

\subsubsection{Data selection}
In our experiment, 20,934 unique syllables and 20,268 unique words from the LMD-full MIDI and reddit MIDI dataset are used for training a skip-gram model to extract embedding vectors of lyrics. As for training the LSTM-GAN model, only paired lyrics-melody sequences in the LMD-full dataset are used. In particular, if a MIDI file has more than 20 notes but less than 40 notes, one 20-note sequence is taken as our data sample; if a MIDI file has more than 40 notes, two 20-note sequences are taken as our data samples. Accordingly, 13,251 sequences each with 20 notes and 265,020 syllable-note pairs are acquired, which are used for training, validation, and testing.

%are made using 20 notes sequences from ``LMD-full'' as both training, validation and testing sets. A sequence of 20 notes is taken from each song from the dataset which has more than 20 notes. Additionally, another sequence of 20 notes is taken from each song from the dataset which has more than 40 notes. This leads to a total of 13,937 sequences, each with 20 notes, i.e. a total of 278,740 syllable-note pairs.

\subsubsection{Parsing MIDI file}
Triplets of music attributes with \{MIDI Number, note duration, rest duration\} are obtained by parsing each MIDI file from the LMD-full dataset which contains English lyrics. The parsing is made as follows:
\begin{itemize}
    \item The beats-per-minute (BPM) value for each MIDI file is extracted.
    \item If a note has a corresponding English syllable, its MIDI Number is extracted. This value is taken as the first of music attributes in our melody representation.
    \item If a note has a corresponding English syllable, its note-on and note-off values are stored.
    \item From the note-on and note-off values, the note duration and rest duration attributes are calculated, using the formula
    \begin{equation}
        x = \phi(t \times \frac{\text{BPM}}{60})
    \end{equation}
     where $x$ is an attribute of note $k$ (either note duration or rest duration), $t$ is a time in seconds ($t = \text{note-off}_\text{k} - \text{note-on}_\text{k}$ to calculate note duration and $t = \text{note-on}_\text{k} - \text{note-off}_\text{k-1}$ to calculate rest duration) and $\phi(\cdot)$ is an operator which constrains the value to the closest value in the set of values used to represent the corresponding attribute.
\end{itemize}

\begin{figure*}[!ht]
    \centering
    \includegraphics[width = 17cm]{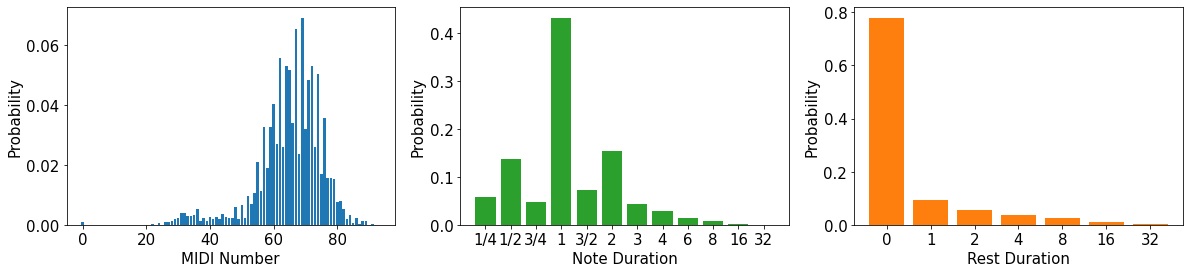}
    \caption{Distribution of music attributes in our dataset.}
    \label{fig:attr}
\end{figure*}

\subsubsection{Note duration}
A note duration means the length of time that a note is played. Note durations frequently appearing in our dataset have been shown in Fig.\ref{fig:attr}, and are summarized in Table \ref{tab:relation1}. When we discretize note lengths of the generated melody, only note durations in Table  \ref{tab:relation1} are considered. In the future, we will try to collect more lyrics which include other note durations.

%\begin{table*}[!ht] \centering
%% \resizebox{\linewidth}{!}
%\begin{tabular}{|c|c|c|c|c|c|c|c|c|c|c|c|c|}
%\hline
%Note duration & 0.25 & 0.5 & 0.75 & 1 & 1.5 & 2 & 3 & 4 & 6 & 8 & 16 & 32 \\
%\hline
%Note   & \Sech{} & \Acht{} & \Acht{}\Pu{} & \Vier{} & \Vier{}\Pu{} & \Halb{} & \Halb\Pu{} & \Ganz{} & \Ganz\Pu{} & $2\times \Ganz{}$ & $4\times %\Ganz{}$ & $8\times \Ganz{}$       \\ \hline
%\end{tabular} %}
%\caption{Relationship between note duration and note}
%\label{tab:configuration}
%$\end{table*}

\begin{table}[!ht] \centering
% \resizebox{\linewidth}{!}
\begin{tabular}{|c|c|c|c|c|c|c|c|c|c|c|c|}
\hline
 0.25  & 0.5 &  0.75   & 1 & 1.5 & 2 & 3 & 4 & 6 & 8 & 16 & 32 \\
\hline
 \Sech{}\!\!\!\!  &  \Acht{}\!\!\!  &  \Acht{}\Pu{}\!\!\!      &  \Vier{} \!\!\! &
 \Vier{}\Pu{}\!\!\!\!\!\!  &  \Halb{} \!\!\!\!\!\!           &  \Halb\Pu{} \!\!\! \!\!\!       &  \Ganz{} \!\!\!\!\!\! &
 \Ganz\Pu{} \!\!\!\!\!\!   &  $\! 2 \!\!\times \! \Ganz{} \!\!$  &  $\!\! 4 \!\! \times \! \Ganz{} \! $  &  $\! 8 \!\! \times \!\! \Ganz{} \!$       \\ \hline

% \! \Sech{} \!      & \! \Acht{} \!           & \! \Acht{}\Pu{} \!      & \! \Vier{} \! &
% \! \Vier{}\Pu{} \! & \! \Halb{} \!           & \! \Halb\Pu{} \!        & \! \Ganz{} \! &
% \! \Ganz\Pu{} \!   & \! $2\times \Ganz{}$ \! & \! $4\times \Ganz{}$ \! & \! $8\times \Ganz{}$ \!      \\ \hline
\end{tabular} %}
\caption{Relationship between note duration and note.}
\label{tab:relation1}
\end{table}

%\begin{itemize}
%    \item 0.25 - \Sech{}
%    \item 0.5 - \Acht{}
%    \item 0.75 - \Acht{}\Pu{}
%    \item 1 - \Vier{}
%    \item 1.5 - \Vier{}\Pu{}
%    \item 2 - \Halb{}
%    \item 3 - \Halb\Pu{}
%    \item 4 - \Ganz{}
%    \item 6 - \Ganz\Pu{}
%    \item 8 - $2\times \Ganz{}$
%    \item 16 - $4\times \Ganz{}$
%    \item 32 - $8\times \Ganz{}$
%\end{itemize}

\subsubsection{Rest}
A rest means how long the silence in a piece of melody will last. The rest values and corresponding rest symbols are shown in Table \ref{tab:relation2}.
\begin{table}[!ht] \centering
% \resizebox{\linewidth}{!}
\begin{tabular}{|c|c|c|c|c|c|c|}
\hline
0 & 1 & 2 & 4 & 8 & 16 & 32 \\
\hline
 No rest \!\!& \ViPa{} \!\!& \HaPa{} (half rest) \!\!& \GaPa{} (whole rest)\!\! & $2 \times \GaPa{}$\!\! & $4\times \GaPa{}$ \!\!\!& $8\times \GaPa{}\!\!$ \\ \hline
\end{tabular} %}
\caption{Relationship between rest values and corresponding symbols.}
\label{tab:relation2}
\end{table}
%\begin{itemize}
 %   \item 0 - No rest
 %  \item 1 - \ViPa{}
 % \item 2 - \HaPa{} (half rest)
 %  \item 4 - \GaPa{} (whole rest)
 %   \item 8 - $2 \times \GaPa{}$
 %   \item 16 - $4\times \GaPa{}$
 %   \item 32 - $8\times \GaPa{}$
%\end{itemize}

\subsubsection{Distribution of music attributes}
The distribution of each attribute in our music dataset is respectively shown in Fig. \ref{fig:attr}, which indicates that most MIDI note numbers range from 60 to 80, quarter note is most frequently played, and $rest=0$ appeared in most cases of melodies.

%\begin{figure}[!ht]
%    \centering
%   \includegraphics[width = 5.5cm]{images/dataset_length_distrib.png}
%  \caption{Dataset Note Durations distribution}
%   \label{fig:dataset_length_distrib}
%\end{figure}

%\begin{figure}[!ht]
%    \centering
%    \includegraphics[width = 5.5cm]{images/dataset_rest_distrib.png}
%    \caption{Dataset Rest Durations distribution}
%    \label{fig:dataset_rest_distrib}
%\end{figure}
\section{Evaluation}
In this section, experimental setup, validation method, and experimental results are introduced to investigate the feasibility of our proposed conditional LSTM-GAN model.

%In this section, the experimental set-up is introduced. First, a short overview of the dataset is given. Then, the Cond-LSTM-GAN configuration is given. The validation process -- which using Maximum Mean Discrepancy (MMD) -- is presented. Lastly, a baseline model against which we will compare our generated sequences is proposed.

\subsection{Experimental setup}
The entire dataset is split with a 0.8/0.1/0.1 proportion between training, validation and testing sets. The model is trained using mini-batch gradient descent for a total of 400 epochs. The learning rate starts at a value of $0.1$, and gradually decreases.
% as shown in Figure \ref{fig:learning_rate}.

%\begin{figure}[!ht]
%    \centering
%    \includegraphics[width = 8cm]{images/learning_rate.png}
%    \caption{Learning rate vs. Epoch}
%    \label{fig:learning_rate}
%\end{figure}

During both validation and testing stages, the sequences of triplet continuous-valued attributes are first constrained to their closest discrete value. The candidate values for the MIDI Numbers are in the range $\{21,\dots,108\}$. In addition, the quantized sequence is checked to see if it belongs to most likely scale, where the MIDI number of the out-of-tune notes is changed to the closest MIDI number in the most likely scale.

\subsection{Validation using MMD}

The validation is made using a Maximum Mean Discrepancy (MMD) \cite{10.1007/978-3-540-75225-7_5} unbiased estimator. Giving two sets of samples, MMD$^2$ takes a value between 0 and 1, indicating how likely the two sets of samples are coming from the same distribution (a value of 0 indicates that the two sets are sampled from the same distribution). Here, a melody sequence of notes is regarded as a set. At the end of each training epoch, the MMD between the generated sequences and the validation sequences is calculated. The weights and biases from the model corresponding to the lowest MMD value are selected.

Let $X_m := \{x_1,\dots,x_m\}$ and $Y_n := \{y_1,\dots, y_n\}$ be two sets of independently and identically distributed (i.i.d) samples from $P_x$ and $P_y$ respectively. Then, an unbiased empirical estimate of MMD$^2$ is given by \cite{2015arXiv151104581B}:

\begin{equation}
\begin{aligned}
    \text{MMD}^2_u &(\mathcal{F}, X_m, Y_n) = \frac{1}{m(m-1)}\sum_{i=1}^m\sum_{j\neq i}^m k(x_i,x_j) \\
    & + \!\! \frac{1}{n(n \!-\! 1)} \!\! \sum_{i=1}^n \!\! \sum_{j \neq i}^n k(y_i, y_j) \! - \! \frac{2}{mn} \!\! \sum_{i=1}^m \!\! \sum_{j=1}^n k(x_i,y_j).
\label{eq:MMD2_unbiased_estimate}
\end{aligned}
\end{equation}
where $\mathcal{F}$ is a reproducing kernel Hilbert space (RKHS), with the kernel function $k(x,x') := \langle \phi(x), \phi(x') \rangle$, and continuous feature mapping $\phi(x) \in \mathcal{F}$ for each $x$.
We used $k(x,x') = \text{exp}(-\|x-x'\|^2/(2\sigma^2))$ as the kernel function, with kernel bandwidth $\sigma$ set such that $\|x-y\|/(2\sigma^2)$ equals 1 when the distance between $x$ and $y$ is the mean distance between points from both datasets $X_m$ and $Y_n$ \cite{median_heuristic}.

\subsection{Comparison methods}
The Random baseline model for the following experiments is inspired by \cite{lee-etal-2019-icomposer}. Melodies of 20 notes are created by randomly sampling the testing set based on the dataset distribution for music attributes (i.e. the distribution shown in Fig. \ref{fig:attr}). Sequences generated by the baseline model are also judged to see if out-of-tune MIDI number needs to be changed. The MIDI numbers are restricted in the set $\{60,\dots,80\}$, meaning that if a MIDI number lower than 60 is sampled from the MIDI numbers distribution, then it takes a value of 60, and similarly a MIDI number higher than 80 is set to 80.

%Besides the random baseline model, we also use a stronger baseline method, which exploits a single LSTM \cite{yu2016seqgan} as the generator.
Besides the Random baseline model, we also use a stronger baseline model trained with the Maximum Likelihood Estimation (MLE) objective which exploits a single-layered LSTM network \cite{yu2016seqgan} as the generator. We refer to it as the MLE baseline model in our experiments.
This MLE baseline is trained by mini-batch gradient descent over the gradient of a log-maximum likelihood estimator as follows:
\begin{equation}
    \hat{l} = - \frac{1}{M\times N} \sum_{i=1}^M\sum_{j=1}^N \log p_{i,j}
\end{equation}
where $M$ is the batch size, $N$ is the sequence length, and $p_{i,j}$ is the estimated probability (i.e. output of the softmax function) with which the $i,j$-th generated note is the same as the $i,j$-th note from the batch taken from the training set. MLE with a single LSTM layer is selected as the baseline according to our experimental results. The configuration for the MLE baseline model is shown in Table \ref{tab:MLE}.
%Since this LSTM generator is trained based on maximum likelihood estimation (MLE), it is abbreviated as MLE, which is called as MLE baseline in following experiments.

\begin{table*}[!ht] \centering
%\resizebox{\linewidth}{!}{
\begin{tabular}{|l|c|}
\hline
        & Generator                                        \\ \hline
Input   & 20 (embedded syllable); 1
(melody-token representing triplet of music attributes)    \\ \hline
Layer 1 & 32, Embedding Layer (for embedding a melody-token) \\ \hline
Layer 2 & 32, LSTM, tanh                                          \\ \hline
Output  & 3397, fully-connected, log-softmax;  3
(predicted triplet of music attributes) \\ \hline
\end{tabular}%}
\caption{Configuration of MLE baseline.}
\label{tab:MLE}
\end{table*}

\subsection{Training stage analysis}
\begin{figure*}[!ht]
    \centering
    \includegraphics[width=14cm]{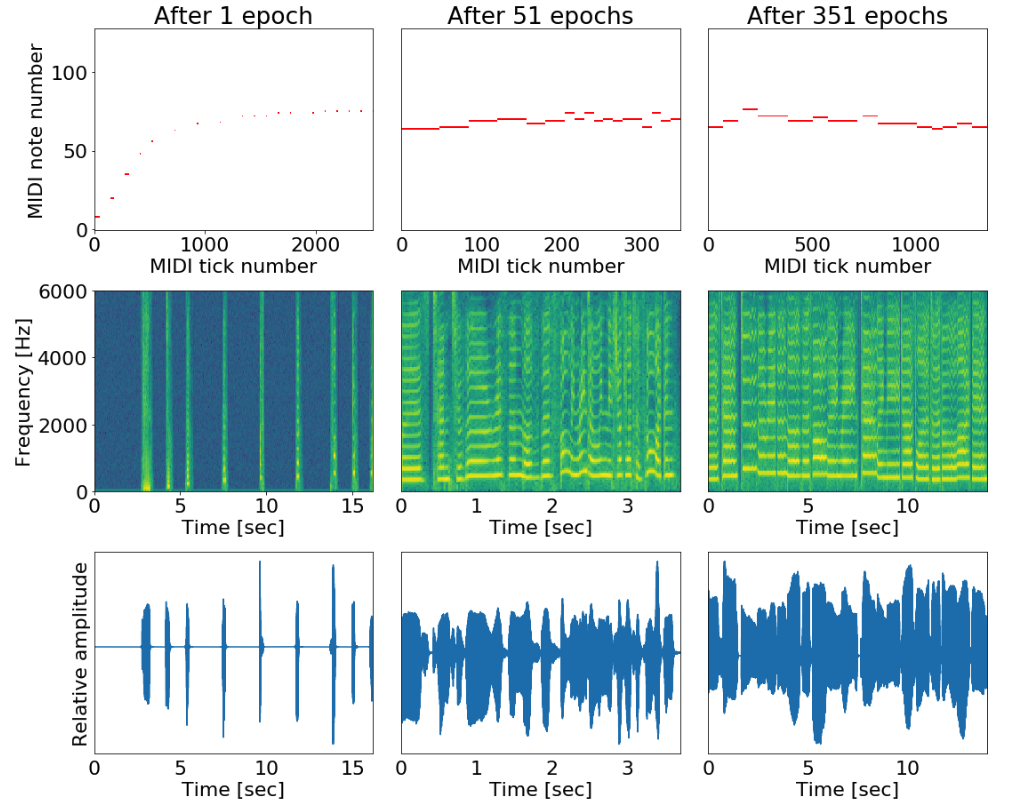}
    \caption{Generated songs by generators trained for 1, 51 and 351 epochs respectively.}
    \label{fig:songs_diff_epochs}
\end{figure*}

\begin{figure}
\subfloat[Model trained for 1 epoch]{\includegraphics[width = 8.9cm]{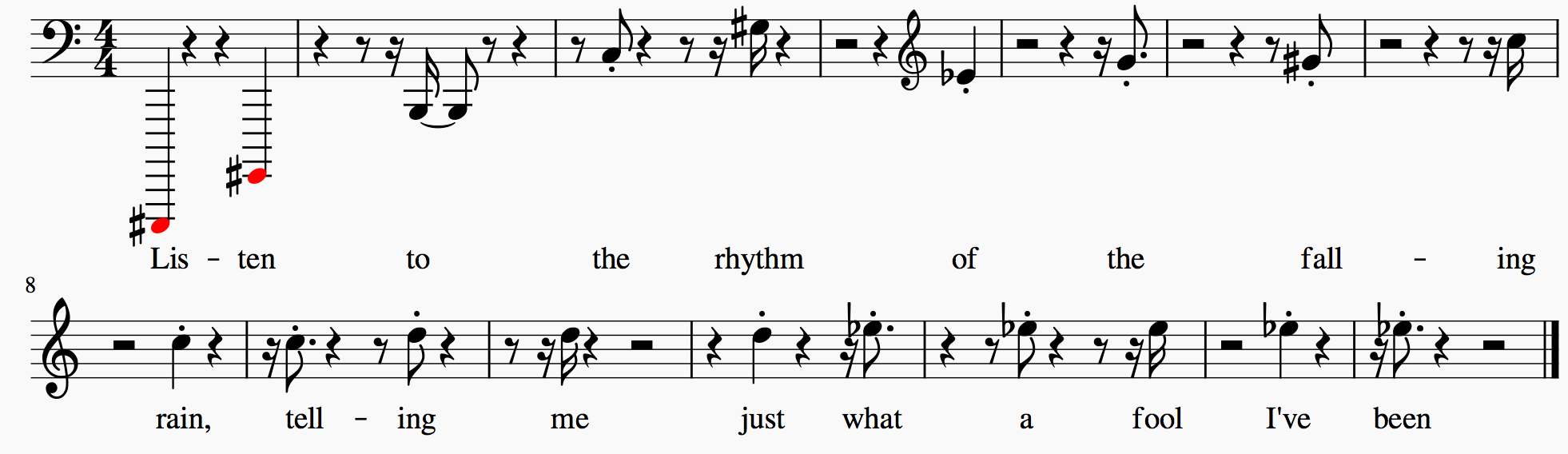}}\\
\subfloat[Model trained for 51 epochs]{\includegraphics[width = 8.9cm]{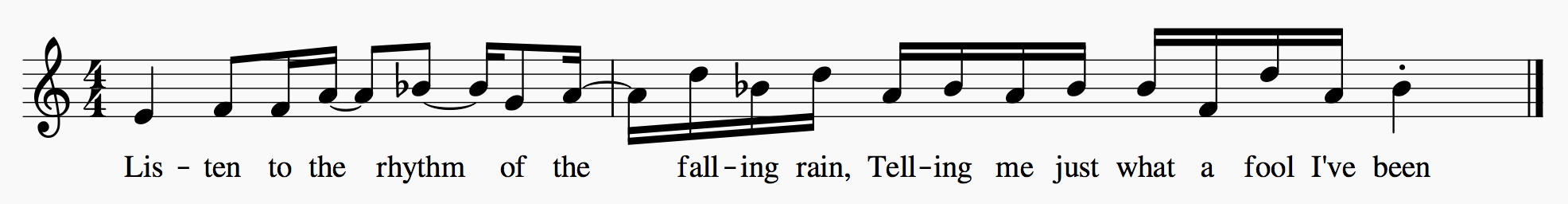}}\\
\subfloat[Model trained for 351 epochs]{\includegraphics[width = 8.9cm]{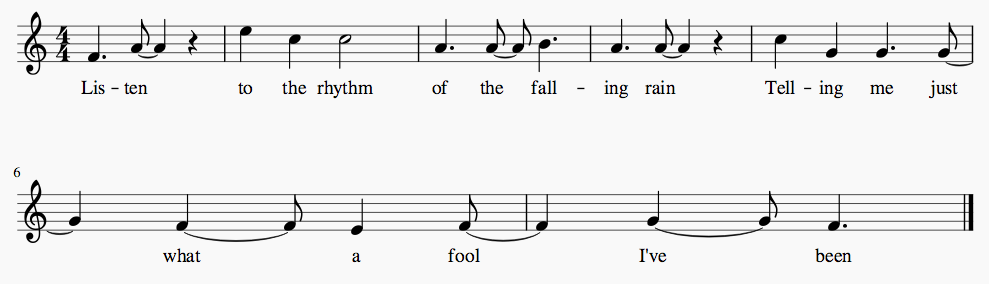}}
\caption{Different sheet music trained for 1, 51 and 351 epochs respectively.}\label{fig:epoch_score}
\end{figure}

The MIDI note number, spectrum, and relative amplitude of generated songs are investigated at 1, 51, and 351 epochs as shown in Fig. \ref{fig:songs_diff_epochs}. The corresponding sheet score with alignment between lyrics and melodies at 1, 51, and 351 epochs are shown in Fig. \ref{fig:epoch_score}. It is obvious that the generated melody gets better when the learning goes deeper by increasing the number of epochs. Additionally, we ask volunteers to listen to the generated songs at different epochs. Their feedback also confirms the effectiveness of our deep conditional LSTM-GAN.

\subsection{Music quantitative evaluation}

In order to investigate if lyrics-conditioned LSTM-GAN can generate melodies that resemble the distribution of training samples, some quantitative measurements are designed to compare the melodies generated by our proposed model, the Random baseline, and MLE baseline, which are shown in the following:
\begin{itemize}
    \item MIDI numbers span: the difference between the highest MIDI number and the lowest one of a sequence.
    \item 3-MIDI-number repetitions: the sum of the ``count of each distinct 3-MIDI occurrence minus 1'' (a melody slice consisting of 3 adjacent notes) throughout the sequence, which is a metric similar to the one used in previous work \cite{Olof}.
    \item 2-MIDI-number repetitions: the sum of the ``count of each distinct 2-MIDI occurrence minus 1'' throughout a sequence.
    \item Number of unique MIDI numbers: a count of how many different MIDI numbers are present in a sequence.
    \item Number of notes without rest: a count of how many rest duration have a value of 0 throughout a sequence.
    \item Average rest value within a song: an average value of the rest duration attribute.
    \item Song length: the sum of all the note duration attributes and all the rest duration attributes of a sequence.
\end{itemize}

\begin{figure*}[!ht]
    \centering
    \includegraphics[width = 16cm]{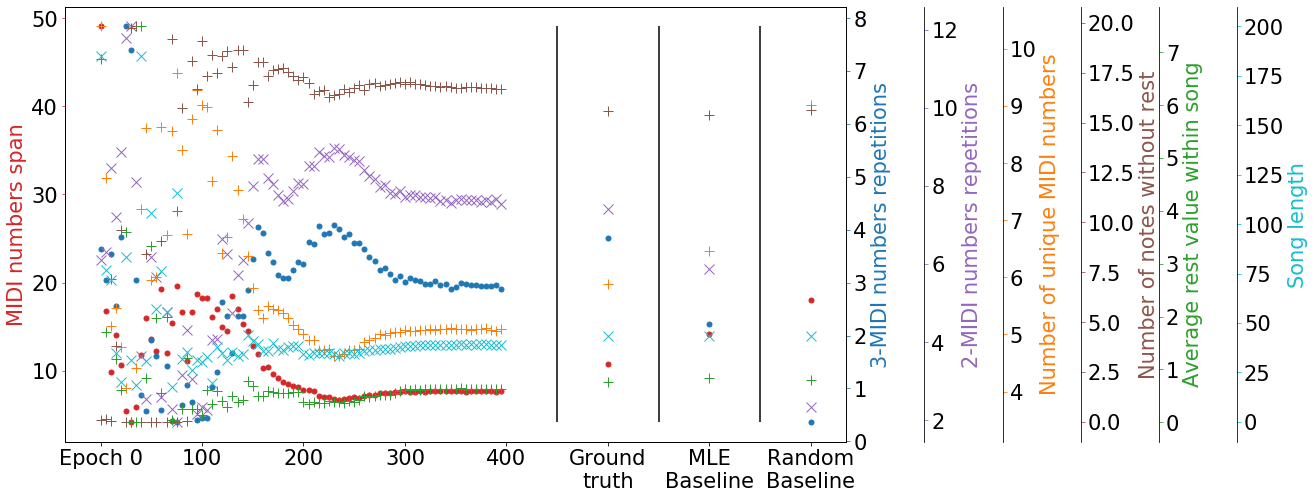}
    \caption{Music-related measurements.}
    \label{fig:music_attr_vs_epoch}
\end{figure*}

Figure \ref{fig:music_attr_vs_epoch} demonstrates the evolution of each of these values averaged over 1,051 generated sequences (one sequence per testing set lyrics).
%For comparison, the value of each measures averaged over the testing set, and for 1394 generated sequences using the baseline model are shown.

For pitch-related attributes, the proposed model outperforms the baseline in every aspect and outperforms MLE in most aspects. The proposed model tends to converge to a value which is relatively close to the ground truth (i.e. the average value from the dataset). However, the 2-MIDI numbers and 3-MIDI numbers repetitions converge to values which are significantly lower than the corresponding measurement over the dataset, but is still much better than that of MLE baseline.

For metrics which are related to temporal attributes (i.e. note duration and rest duration), the baselines are closer to the ground truth value. This is expected, since these metrics are nothing but an average of attributes which the baseline samples from the ground truth distribution. Hence, these values tend to the ground truth value as the number of generated examples increases. Table \ref{tab:in-songs_results} shows the numerical values of the results.

We also investigated scale-consistency of generated melodies. The mean accuracy of scale-consistency for the conditional LSTM-GAN model is 48.6\%. In contrast, for the MLE baseline and Random baseline models, it is 47.3\% and 46.6\% respectively. The mean accuracy of scale-consistency is not high due to two main reasons: i) the mapping between lyrics and melody is not unique, i.e., for a given lyrics there can be multiple melodies belonging to different standard scales. ii) some notes overlap between different standard scales, e.g., D,E,F,G,A appear in both $\mathcal{C}_{major}$ and $\mathcal{D}_{natural\ minor}$.

\begin{table*}[!ht]\centering
\begin{tabular}{|l|c|c|c|c|}
\hline
                               & Ground truth      & Conditional LSTM-GAN         & MLE baseline        & Random baseline \\ \hline
%                               & Ground truth & Cond LSTM-GAN  & MLE baseline & Random baseline\\ \hline
MIDI numbers span              & 10.8        & 7.7             & 14.2 & 18.1 \\ \hline
3-MIDI numbers repetitions     & 3.8         & 2.9             & 2.2 & 0.4\\ \hline
2-MIDI numbers repetitions     & 7.4        & 7.7            &5.9 & 2.3\\ \hline
Number of Unique MIDI          & 5.9        & 5.1             & 6.5 & 9.0\\ \hline
Number of notes without rest   & 15.6       & 16.7            & 15.4 & 15.6 \\ \hline
Average rest value within song & 0.8         & 0.6             & 0.8 & 0.8 \\ \hline
Song length                    & 43.3        & 39.2           & 43.4  & 43.6\\ \hline
%\textcolor{red} {\# matched scale consistency}   & \textcolor{red}{1051}        & \textcolor{red}{511}       & \textcolor{red}{497}  & \textcolor{red}{490}\\ \hline
\end{tabular}
\caption{In-songs attributes metrics evaluation.}
\label{tab:in-songs_results}
\end{table*}

\begin{figure*}[!ht]
    \centering
    \includegraphics[width = 17cm]{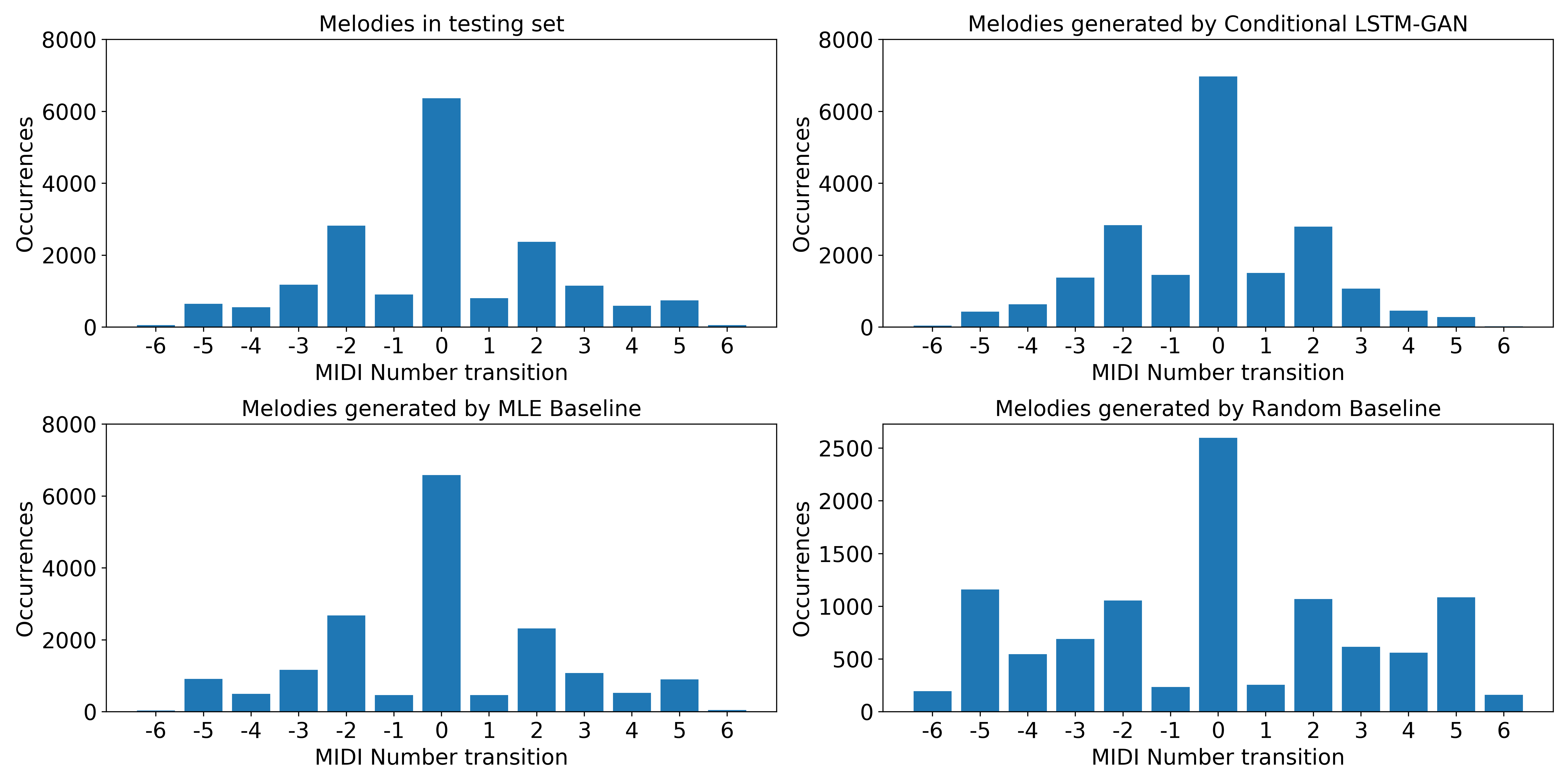}
    \caption{Distribution of transitions.}
    \label{fig:transitions_dist}
\end{figure*}

Another important attribute in music is the distribution of the transitions between MIDI numbers. Figure \ref{fig:transitions_dist} shows the distributions of the transitions for the melodies generated by our model, the MLE and Random baseline models and the testing set. The melody generated by the proposed model approximates well the distribution of the human composed music. Although the testing set melody has a slightly higher transition from a note to a lower-pitched one, the melody generated by our model has more transitions from a note to a higher-pitched one.

To compare the qualities of generated melodies, standard BLEU scores as evaluation metrics \footnote{https://en.wikipedia.org/wiki/BLEU} for conditional LSTM-GAN, MLE baseline, and Random baseline are respectively calculated, which are shown in Table \ref{tab:BLEU}. From this table, we can find that our conditional LSTM-GAN model achieves the highest BLEU scores compared with the other two baseline methods. This also concludes that our method can generate melodies of relatively higher qualities.

\begin{table}[!ht]
\resizebox{\linewidth}{!}{
\begin{tabular}{|l|c|c|c|}
\hline
   Scores              & BLEU-2            & BLEU-3           & BLEU-4            \\ \hline
Random baseline        & 0.499             & 0.157            &  0.036             \\ \hline
MLE baseline           & 0.659             & 0331             & 0.137              \\ \hline
Conditional LSTM-GAN   & 0.735    & 0.460   & 0.230         \\ \hline
\end{tabular}}
\caption{BLEU scores for Random and MLE baseline methods and proposed conditional LSTM-GAN.}
\label{tab:BLEU}
\end{table}

\subsection{Effect of lyrics conditioning}

\subsubsection{Distribution of MIDI numbers}
\begin{figure}[!ht]
    \centering
    \includegraphics[width=6.0cm]{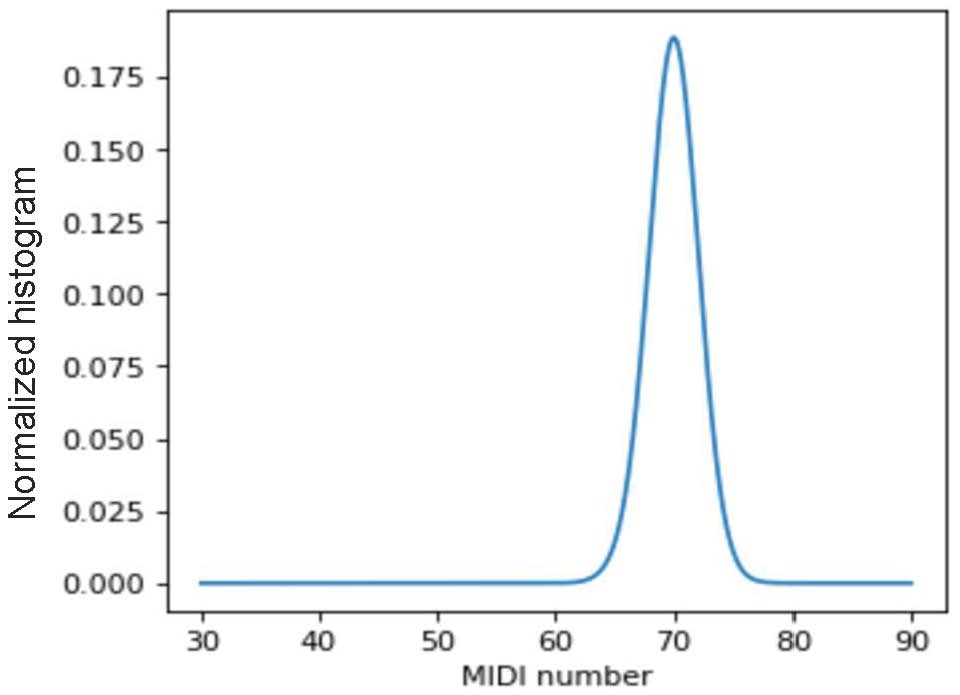}
    \caption{Estimated distribution of generated MIDI numbers (no conditioning on lyrics).}
    \label{fig:nocon}
\end{figure}

\begin{figure*}[!ht]
    \centering
    \includegraphics[width=12.8cm]{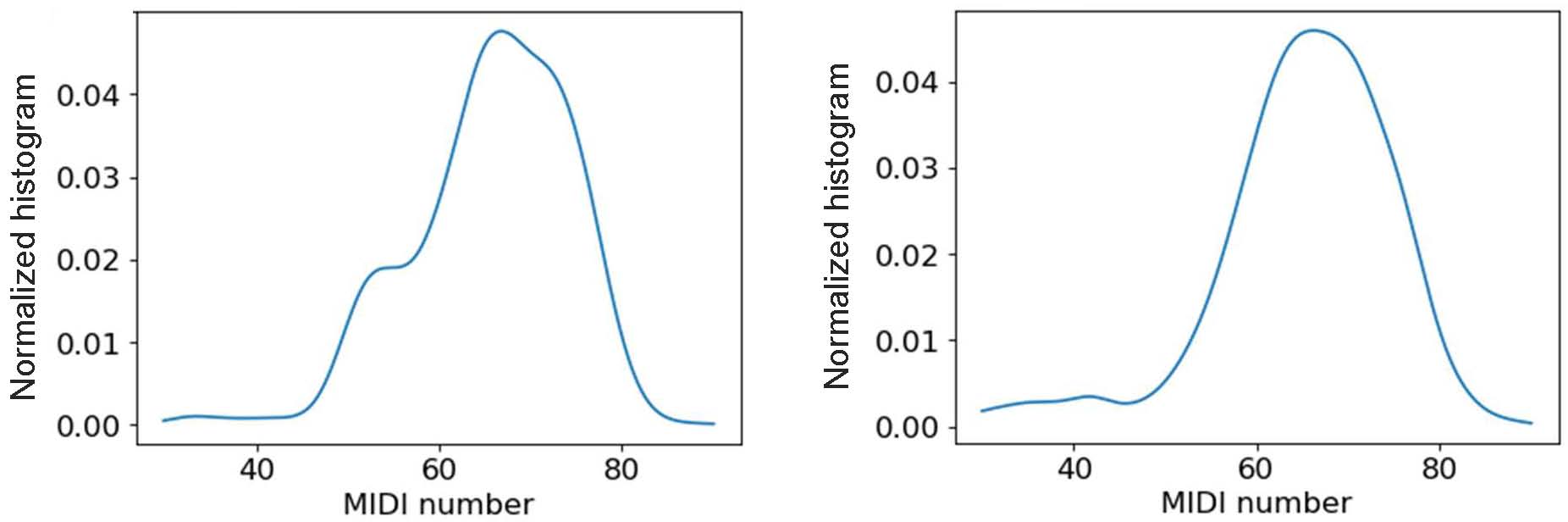}
    \caption{Left: estimated distribution of generated MIDI numbers conditioned on lyrics. Right: estimated distribution of ground truth MIDI numbers.}
    \label{fig:distrib_pitch_generated}
\end{figure*}

The experiments show that the proposed LSTM-GAN model without conditioning on lyrics leads to a model which generates a narrower-band distribution of MIDI number as shown in Fig. \ref{fig:nocon}. Here are the MIDI number distributions, which are smoothed by using kernel density estimation. It is obvious in Fig. \ref{fig:distrib_pitch_generated} that the estimated distribution of generated MIDI numbers conditioned on lyrics is more consistent with the distribution of ground truth MIDI numbers.

\begin{figure}
    \centering
    \includegraphics[width=0.4\textwidth]{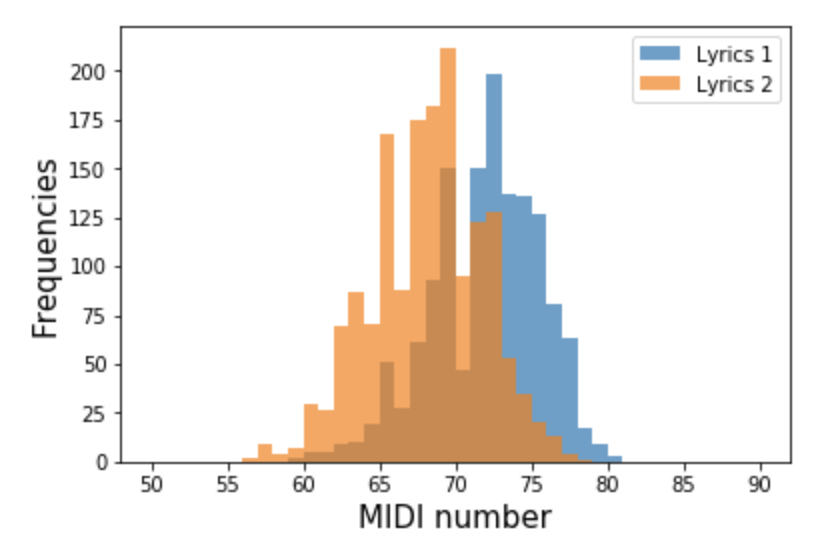}
    \caption{Estimated distributions of MIDI numbers generated by two different lyrics. Lyrics 1 are taken from the song \textit{It Must Have Been Love (Christmas for the Broken Hearted)} by Roxette (written by Per Gessle). Lyrics 2 are taken from \textit{Love's Divine} by Seal (written by Mark Batson and Sealhenry Samuel).}
    \label{fig:lyrics_midi_numbers}
\end{figure}

An example of the influence of lyrics on the generated MIDI numbers is shown in Fig. \ref{fig:lyrics_midi_numbers}. Given two kinds of lyrics, 1,000 songs are generated for each lyrics, and the distribution of the generated MIDI numbers is estimated. In our example, the second lyrics leads to songs with lower MIDI numbers than the first one, which helps to deliver some semantic information latent in the lyrics.

\subsubsection{Note duration and rest duration}
In this experiment, an evaluation method to seeing if the lyrics conditioning has an effect on the generated note duration and rest duration is presented. The following focuses on note duration attribute, but the same is valid for rest duration as well.

Let $D \in \mathbb{R}^{N\times M}$ be a matrix composed of N note duration (or note rest) sequences of M note durations, where each sequence is taken from a different song from a N songs dataset used to train a generator G. Therefore, $D_{i,j}$ is the $j$-th note duration of the $i$-th song. Let $G \in \mathbb{R}^{N\times M}$ be a matrix composed of note duration sequences generated by G by feeding the syllables sequences corresponding to each row of $D$. This means that $G_{i,j}$ is the $j$-th note duration of the sequence generated by G when the syllables corresponding to $\mathbf{d}_i$ are fed to it, where $\mathbf{d}_i$ denotes the $i$-th row of $D$.

Let $\text{randrow}(\cdot)$ be an operator which randomizes the order of the row of a matrix. Therefore $D_{\text{rs}} = \text{randrow}(D)$ can be seen as a matrix with correct in-sequence note duration order, but wrong song order when compared to $D$. $D_{\text{rn}} = \big(\text{randrow}(D^T)\big)^T$ can be seen as a matrix for which the song order is the same as $D$'s, but the note duration sequences are randomized. Finally, $D_{\text{rns}} = \text{randrow}(D_{\text{rn}})$ can be seen as a matrix in which both the song and note order are randomized when compared to $D$. The subscripts $_{\text{rs}}$, $_{\text{rn}}$ and $_{\text{rns}}$ denote ``random songs", ``random notes", and ``random notes + songs" respectively. Since $\text{randrow}(\cdot)$ is a random operator, $D_{\text{rs}}$, $D_{\text{rn}}$, $D_{\text{rns}}$ are matrices of random variables (random matrices).

In this experiment, $d = \frac{1}{NM}\|D-G\|_F$ (which is a real value) is compared to the distribution of the random variables $d_{\text{rs}} = \frac{1}{NM}\|D_{\text{rs}}-G\|_F$, $d_{\text{rn}} = \frac{1}{NM}\|D_{\text{rn}}-G\|_F$ and $d_{\text{rns}} = \frac{1}{NM}\|D_{\text{rns}}-G\|_F$, with $N = 1,051$ (number of songs in testing set) and $M=20$. The experiment is made on the testing set.

\begin{figure}[!ht]
    \centering
    \includegraphics[width=8.8cm]{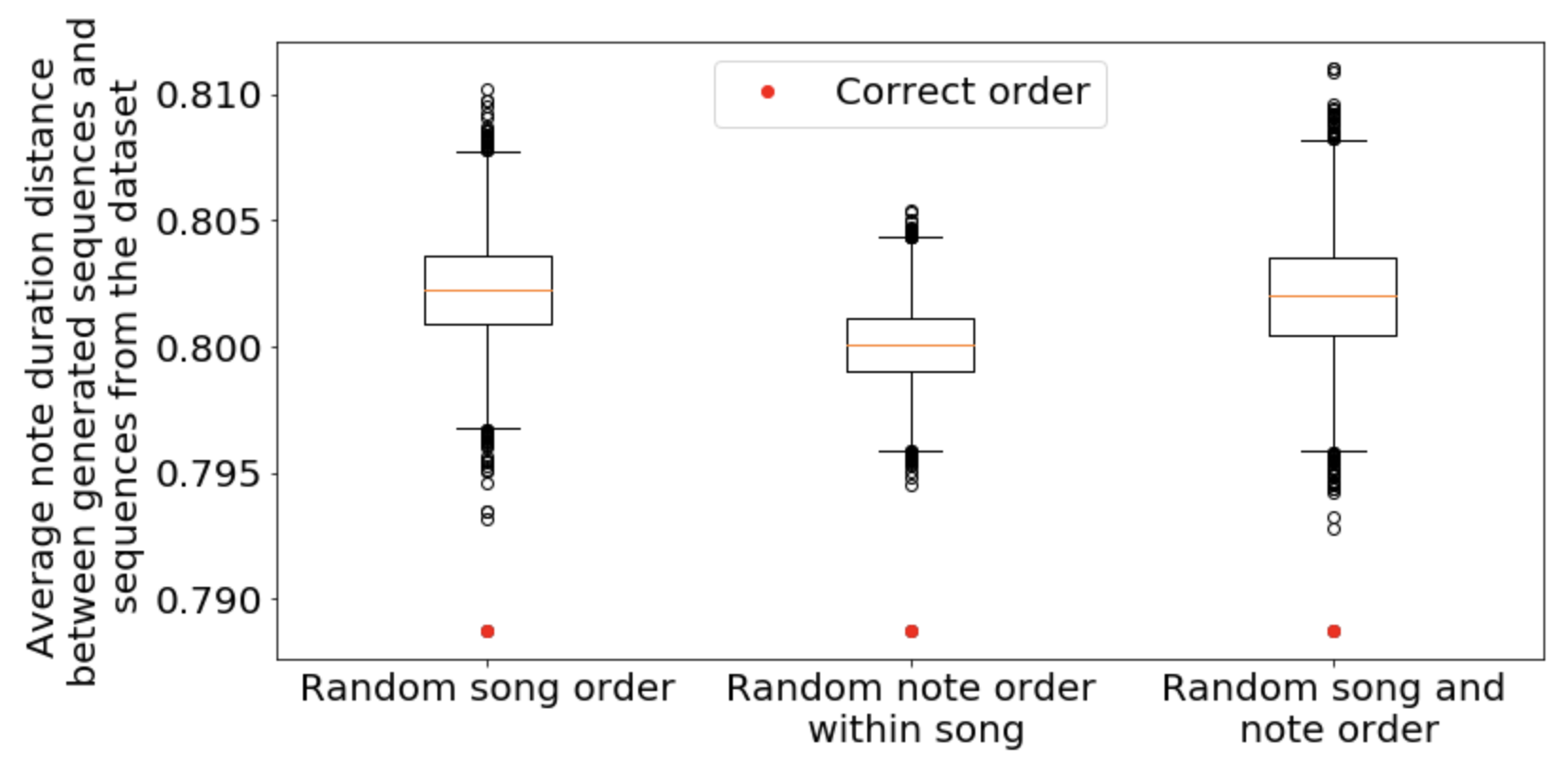}
    \caption{Boxplots of the distributions of $d_{\text{rs}}$, $d_{\text{rn}}$ and $d_{\text{rns}}$ (For the note duration attribute). $d = 0.788$ is highlighted in red in each boxplot. Mean values are $\mu_{\text{rs}} = 0.802$, $\mu_{\text{rn}} = 0.801$ and $\mu_{\text{rns}} = 0.802$ respectively.}
    \label{fig:lengths_distance_boxplot}
\end{figure}

\begin{figure}[!ht]
    \centering
    \includegraphics[width=8.8cm]{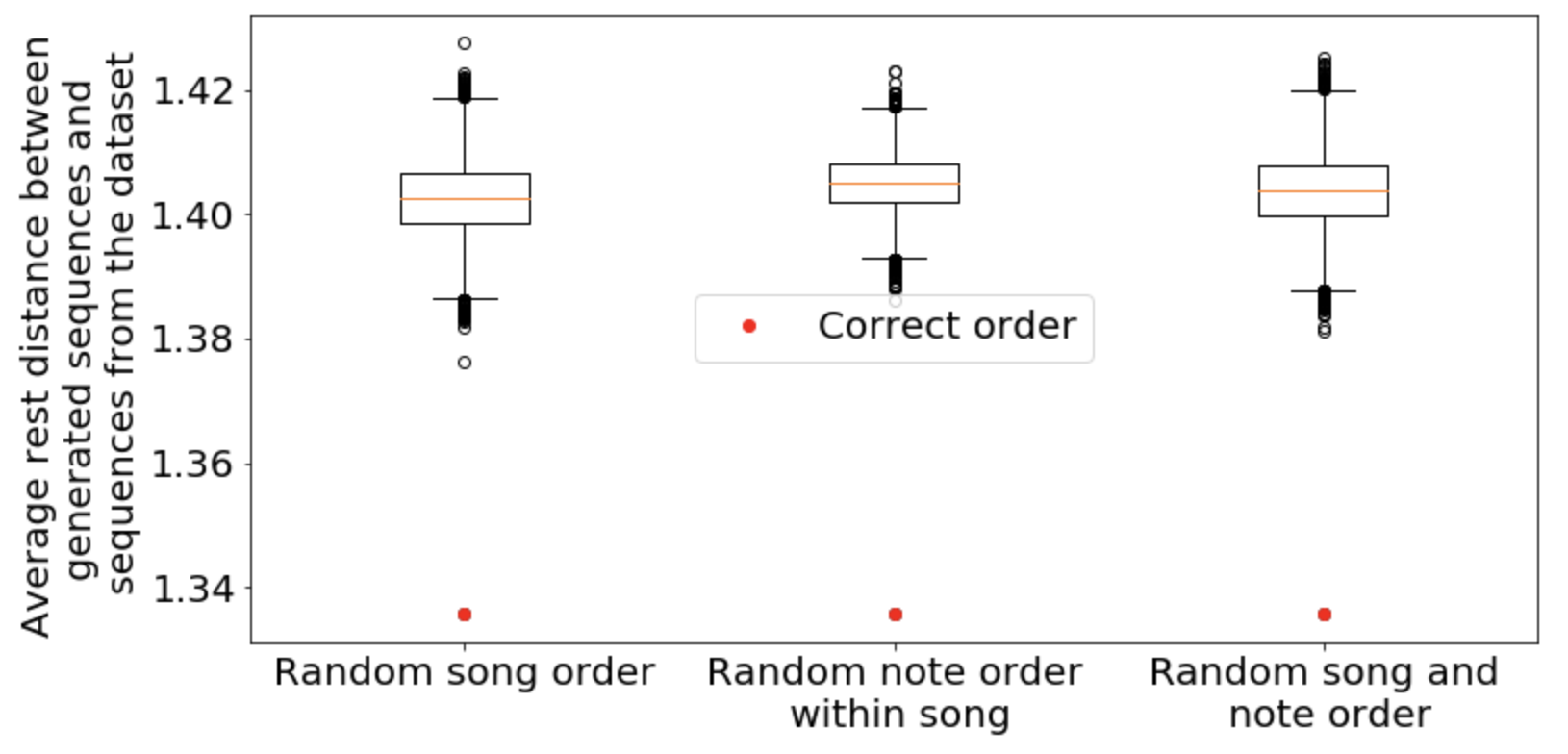}
    \caption{Boxplots of the distributions of $d_{\text{rs}}$, $d_{\text{rn}}$ and $d_{\text{rns}}$ (for the rest duration attribute). $d = 1.336$ is highlighted in red in each boxplot. Mean values are $\mu_{\text{rs}} = 1.404$, $\mu_{\text{rn}} = 1.407$ and $\mu_{\text{rns}} = 1.404$ respectively.}
    \label{fig:rests_distance_boxplot}
\end{figure}

Results are shown in Fig. \ref{fig:lengths_distance_boxplot} (note duration), and Fig. \ref{fig:rests_distance_boxplot} (rest duration). The three distributions are estimated using 10,000 samples for each random variable. In each case, $d$ is statistically lower
than the mean value, indicating $G$ learned useful correlation between syllable embeddings and note/rest durations.

\subsection{Subjective evaluation}
\begin{figure*}[!ht]
    \centering
    \includegraphics[width = 17cm]{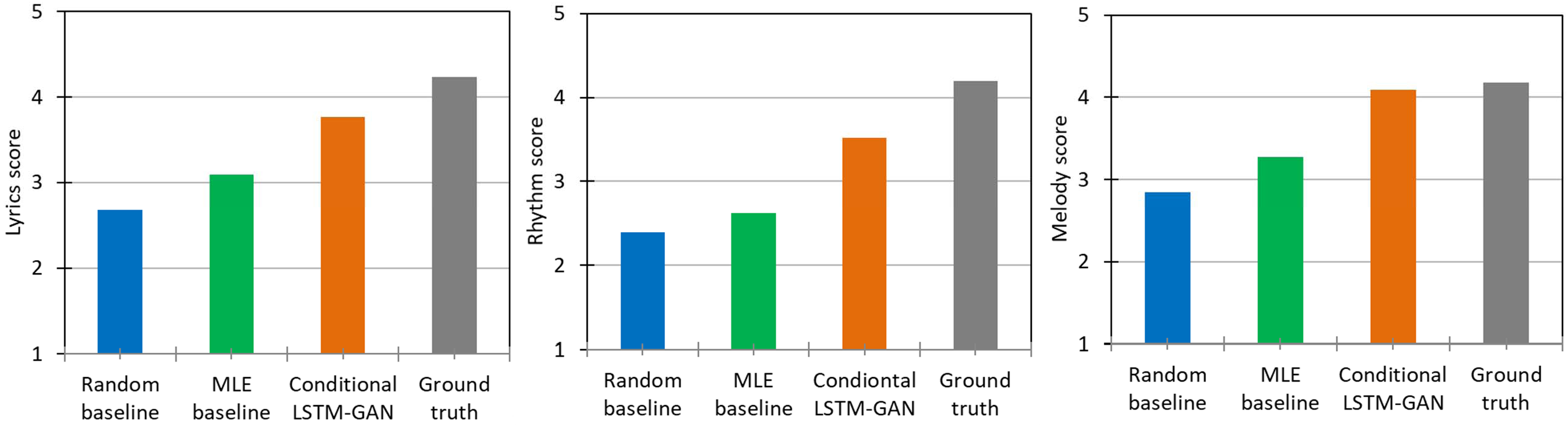}
    \caption{Subjective evaluation results.}
    \label{fig:subj_eval_results}
\end{figure*}

%Sythesizer V\footnote{https://synthesizerv.com/en/}
4 different lyrics are randomly selected from the ground truth dataset. Accordingly, 16 melodies are obtained by using Random baseline, MLE baseline, our model, and ground truth. These melodies can be downloaded from the link \footnote{\url{https://drive.google.com/file/d/1ugOwfBsURax1VQ4jHmI8P3ldE5xdDj0l/view?usp=sharing}}.

All melodies are sung by a female voice produced by synthesizer V \cite{synt}. 4 male and 3 female subjects without knowing our research and any musical knowledge were invited to listen to these 16 melodies, where each melody with around 20 seconds is played 2 times in a random order. Three questions as metrics are used for evaluation: how about the entire melody? how about the rhythm? and does the melody fit the lyrics well? The subjects are asked to give a score from 1 to 5 (where 1 corresponds to ``very bad", 2 to ``bad", 3 to ``OK", 4 to ``good" and 5 to ``very good").

The first run is taken to enable subjects to get used to the type of melodies they were listening to. The scores of evaluation metrics are respectively averaged based on listening results of Random baseline, MLE baseline, our model, and ground truth on the second run.

Evaluation results are shown in Figure \ref{fig:subj_eval_results}. It is obvious that the melodies generated by the proposed model are closer to the ones composed by humans than the baseline in each metric. This also verifies the proposed conditional LSTM-GAN can predict the plausible sequential alignment relationship between syllables and notes. The feedback from subjects indicates that relatively low scores of melody evaluation are generated which might be due to the limited capability of the synthesizer for high pitches. From these results of all three metrics, we also can find there still are the gaps between melodies generated by our model and ones from human composition, which tells us there is much space we can investigate to improve capability of neural melody generation.

\section{Conclusion and future work}
Melody generation from lyrics in music and AI is still unexplored well. Making use of deep learning techniques for melody generation is a very interesting research area, with the aim of understanding music creative activities of human. Several contributions are done in this work: i) the largest paired English lyrics-melody dataset is built to facilitate the learning of alignment relationship between lyrics and melody. This dataset is very useful for the area of melody generation. ii) a skip-gram model is trained to exact lyrics embedding vectors, which can be taken as a lyrics2vec model for English lyrics feature extraction. iii) Conditional LSTM-GAN is proposed to model sequential alignment relationship between lyrics and melody, followed by a tuning scheme that has the capability of constraining a continuous-valued sequence to the closest in-tune discrete-valued sequence. iv) Evaluation method of melody generation is suggested to demonstrate the effectiveness of our proposed deep generative model.

The continuous-valued sequence generated by the generator needs to be constrained to the underlying musical representation of discrete-valued MIDI attributes. Due to the quantization error, the generated note could be associated to an improper duration, which would destroy the rhythm. But this probability could be low after applying LSTM in the melody generation. In the future, we plan to apply Gumbel-Softmax relaxation technique \cite{jang2016} to train a lyrics-conditioned LSTM-GAN for the discrete generation of music attributes.

In addition, several interesting future works will be investigated as follow: i) How to compose a melody with the sketch of uncomplete lyrics, ii) how to compose a polyphonic melody with lyrics, and iii) how to predict lyrics when given a melody as a condition.

\bibliographystyle{IEEEtran}
\bibliography{IEEEabrv,biblio}

\end{document}